\documentclass[10pt,twocolumn,letterpaper]{article}

\usepackage[pagenumbers]{cvpr} 

\usepackage{graphicx}
\usepackage{amsmath}
\usepackage{amssymb}
\usepackage{booktabs}

\usepackage[pagebackref,breaklinks,colorlinks]{hyperref}

\usepackage{multirow}
\usepackage{enumitem}

\usepackage[capitalize]{cleveref}
\crefname{section}{Sec.}{Secs.}
\Crefname{section}{Section}{Sections}
\Crefname{table}{Table}{Tables}
\crefname{table}{Tab.}{Tabs.}

\def\cvprPaperID{3132} 
\def\confName{CVPR}
\def\confYear{2022}

\begin{document}

\title{ProposalCLIP: Unsupervised Open-Category Object Proposal Generation 
	
	via Exploiting CLIP Cues}

\author{Hengcan Shi,\quad  Munawar Hayat,\quad  Yicheng Wu,\quad  Jianfei Cai\\
	Monash University\\
	{\tt\small \{hengcan.shi, munawar.hayat, yicheng.wu, jianfei.cai\}@monash.edu}}
\maketitle

\begin{abstract}
   Object proposal generation is an important and fundamental task in computer vision. In this paper, we propose ProposalCLIP, a method towards unsupervised open-category object proposal generation. Unlike previous works which require a large number of bounding box annotations and/or can only generate proposals for limited object categories, our ProposalCLIP is able to predict proposals for a large variety of object categories without annotations, by exploiting CLIP (contrastive language-image pre-training) cues. Firstly, we analyze CLIP for unsupervised open-category proposal generation and design an objectness score based on our empirical analysis on proposal selection. Secondly, a graph-based merging module is proposed to solve the limitations of CLIP cues and merge fragmented proposals. Finally, we present a proposal regression module that extracts pseudo labels based on CLIP cues and trains a lightweight network to further refine proposals. Extensive experiments on PASCAL VOC, COCO and Visual Genome datasets show that our ProposalCLIP can better generate proposals than previous state-of-the-art methods. Our ProposalCLIP also shows benefits for downstream tasks, such as unsupervised object detection.
\end{abstract}

\section{Introduction}
\label{sec:intro}
Object proposal generation aims to predict a number of category-agnostic bounding box proposals for all objects in an image. It serves as a fundamental and crucial step towards many higher-level tasks, such as object detection \cite{xie2021detco, liu2021swin, girshick2015fast, ren2015faster}, object segmentation \cite{DBLP:conf/iccv/HeGDG17, wang2021end, chen2020blendmask} and image captioning \cite{mahajan2020diverse, Li2019Know}. How to effectively generate as few as possible proposals to cover all objects is the key challenge in object proposal generation.

Traditional proposal generation methods \cite{alexe2012measuring, uijlings2013selective, zitnick2014edge, cheng2014bing, zhang2017sequential} often utilizes low-level cues (e.g., color, texture, gradient and/or edge) to select proposals from sliding window boxes. In recent years, deep-learning-based methods \cite{liu2020refinedbox, kuo2015deepbox, ren2015faster, zitnick2014edge} take high-level semantics from CNNs or Transformers as cues to select or regress proposals. Although these deep-learning-based methods significantly improve the proposal generation performance, they require a huge number of bounding box annotations for training, which is very labor-intensive, especially for large-scale datasets. Meanwhile, due to the significant annotation effort required, only objects of a limited number of categories can be labeled. Thus, these supervised methods can only generate proposals for limited object categories. However, real-world applications such as object retrieval \cite{Hu2016Natural, bai2021connecting}, image captioning \cite{mahajan2020diverse, Li2019Know} and referring grounding \cite{yu2018mattnet, Hu2016Natural} usually require object proposals of diverse categories.

\begin{figure}
	\centerline{\includegraphics[scale=0.55]{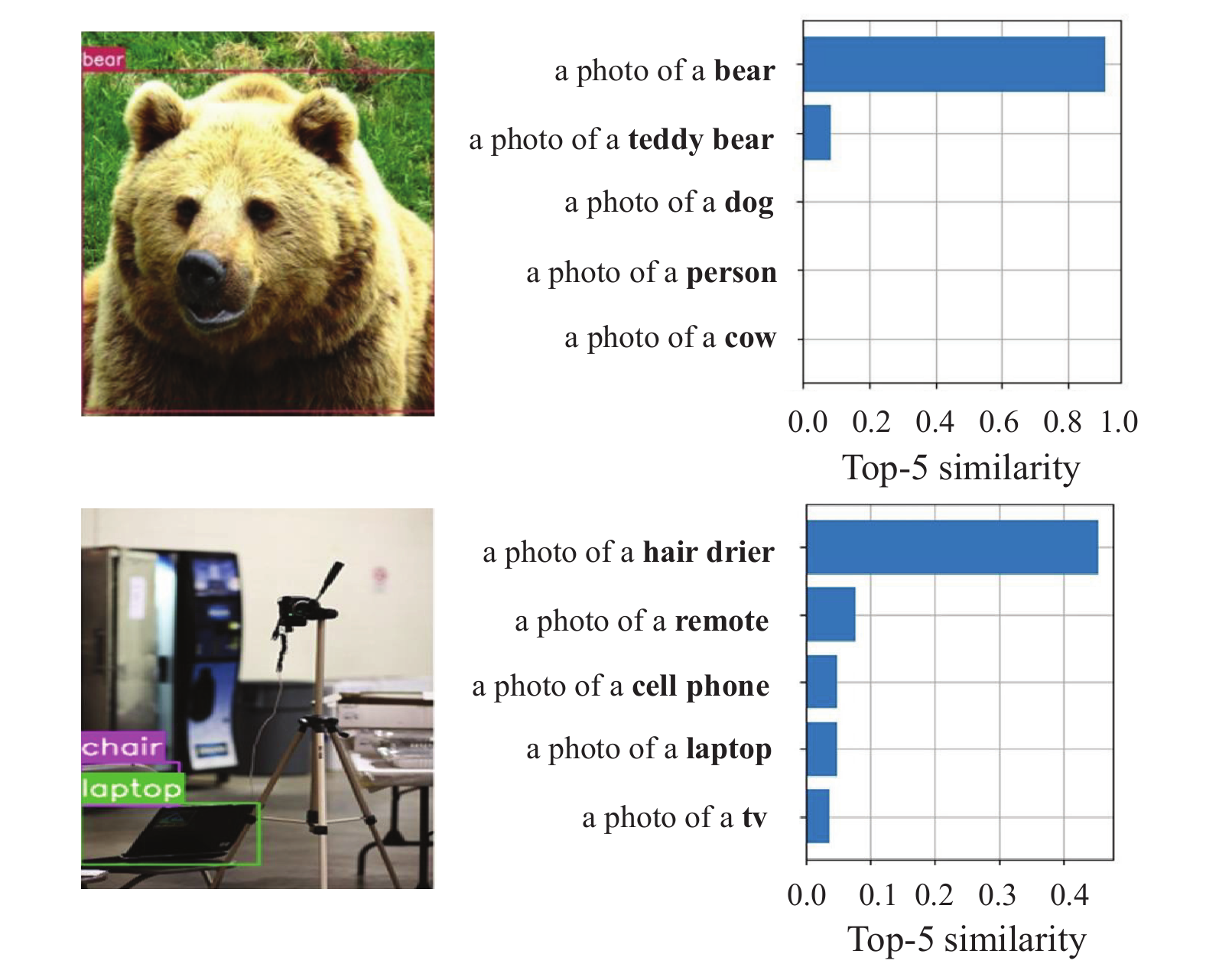}}
	\caption{Examples of CLIP \cite{radford2021learning} image-text matching results. This pre-trained model well recognizes objects of open categories. However, it can only highlight a single object in an image. For example, the ``chair'' and ``laptop'' are not recognized in the bottom image.}
	\label{fig_introduction}
\end{figure}

Some recent efforts \cite{joseph2021towards, zareian2021open, vo2020toward, simeoni2021localizing} aim to address these challenges. ORE \cite{joseph2021towards} and OVR-CNN \cite{zareian2021open} leverage incremental learning and image caption supervisions to recognize additional object categories. However, these approaches also need a mass of bounding box and image caption annotations. Without the human-intensive annotations, these methods fail to perform well.
LOST \cite{simeoni2021localizing} and rOSD \cite{vo2020toward} propose unsupervised deep-learning-based proposal generation approaches, which predict proposals based on class activation maps (CAMs) and attention maps from pre-trained classification networks. These unsupervised methods avoid the box annotation, but they are only able to recognize limited object categories. In addition, although CAMs/attention maps activate some salient regions, there are many objects in non-activated regions. As a result, proposals generated by these methods can only cover parts of objects, as shown in Tables \ref{tab_coco} and \ref{tab_vg}.

In this paper, we propose a novel method, called ProposalCLIP, towards unsupervised and open-category object proposal generation. Our method can generate a variety of proposals of different object categories without requiring expensive bounding box annotations, by leveraging the pre-trained image-text matching model, CLIP (contrastive language-image pre-training) \cite{radford2021learning}. We exploit CLIP \cite{radford2021learning} features because it is trained on millions of image-language pairs from web, and thus have potential to generalize to various object categories, as shown in Fig. \ref{fig_introduction}. Nevertheless, CLIP \cite{radford2021learning} cannot be directly used in object proposal generation, because it is only trained to recognize single-object images and cannot well handle multi-object images. For example, in the second image in Fig. \ref{fig_introduction}, it only ignores both the ``chair'' and ``laptop'' objects. Thus, it is non-trivial to apply CLIP for our task. In our ProposalCLIP, we first analyze CLIP features and build an objectness score based on our analysis for proposal generation. In addition, we design a graph-based proposal merging model, which exploits CLIP features to effectively combine different proposals. We also extract pseudo labels based on CLIP cues to train a box regression model to further improve our proposals. We conduct experiments on three common datasets, PASCAL VOC (20 classes), COCO (80 classes) and Visual Genome (1,600 classes). The experimental results demonstrate the effectiveness of our proposed method.

Our major contributions can be summarized as follows: (1) We propose a novel method that can effectively generate proposals for open categories in real world, without requiring annotations. (2) To the best of our knowledge, this is the first study to analyze and exploit CLIP cues as prior knowledge for object proposal generation. We analyze the CLIP for proposal generation and design a CLIP proposal selection model, a graph-based proposal merging model as well as a proposal regression model to further refine and tailor CLIP cues. (3) Extensive experimental results show that our proposed framework obtains significant improvements on three popular datasets and has benefits for downstream tasks.

\section{Related Work}
\textbf{Supervised Object Proposal Generation.} Fully-supervised object proposal generation methods use bounding box annotations to train models and select proposals from initial bounding boxes. BING \cite{cheng2014bing} used normed image gradients (NG) as the cue and trained a support vector machine (SVM) to select proposals. BING++ \cite{zhang2017sequential} further incorporated edges and segments to improve the proposal
localization quality. These methods are based on low- and mid-level cues. Many deep-learning-based methods have been developed to explore higher-level cues for the proposal generation task. DeepBox \cite{kuo2015deepbox} designed a four-layer CNN to re-rank initial proposals generated by Edge Boxes \cite{zitnick2014edge}. Faster RCNN \cite{ren2015faster} built a region proposal network (RPN) including a classifier and a box regressor to select and correct bounding boxes generated by anchors. RFP-Net \cite{jiao2020rfp} and Refinedbox \cite{liu2020refinedbox} modified RPN. RFP-Net \cite{jiao2020rfp} used receptive fields (RFs) to generate initial boxes to remove many hyper-parameters of anchor boxes in RPN. Refinedbox \cite{liu2020refinedbox} replaced the classifier in RPN with a ranking model to re-rank bounding boxes from Edge Boxes \cite{zitnick2014edge}. A number of methods \cite{DBLP:conf/cvpr/RedmonDGF16, liu2016ssd, carion2020end} also trained deep networks to directly regress object proposals. Although these fully-supervised methods achieve high-quality proposals, they need a huge set of bounding box annotations for training. 

To reduce the requirement of human annotations, weakly-supervised methods \cite{zhou2016learning, zhu2017soft, selvaraju2017grad, tang2018weakly, cheng2020high} only use image-level labels rather than bounding box annotations. They usually leverages image-level labels to train a classification network and generate class activation maps (CAMs) from the trained model. Then, they extract proposals from these CAMs. Some of weakly-supervised methods \cite{zhu2017soft, tang2018weakly, cheng2020high} also select high-confidence proposals as pseudo labels to train an object proposal generation model in a fully-supervised manner to improve the accuracy. However, weakly-supervised methods still require costly image classification annotations and human annotators. In addition, both fully- and weakly- supervised method can only generate proposals for a limited number of object categories.

\textbf{Unsupervised Object Proposal Generation.} To avoid these limitations of supervised object proposal generation, unsupervised methods have attracted increasing research interests \cite{uijlings2013selective, zitnick2014edge} in recent years, which does not need annotations on the target dataset. Early methods such as Selective Search \cite{uijlings2013selective} and Edge Boxes \cite{zitnick2014edge} employed color, texture or edge cues to predict proposals. These methods contain no training process and thus avoid human annotations. However, these methods are only able to leverage low-level information. High-level information such as deep learning features are hard to be used in unsupervised settings,, because deep learning requires training data and annotations. To solve this problem, Detco \cite{xie2021detco} proposed to use contrastive learning to train the deep learning model in a self-supervised manner. Nevertheless, the model trained by contrastive learning can only extract features. A fully-supervised detector is still needed for bounding boxes prediction. Inspired by weakly supervised techniques, some recent methods \cite{wei2019unsupervised, vo2020toward, simeoni2021localizing} predict proposals from CAMs or attention maps provided by pre-trained classification models. Wei \emph{et al.} \cite{wei2019unsupervised}, Vo \emph{et al.} \cite{vo2020toward} and Simeoni \emph{et al.} \cite{simeoni2021localizing} employed PCA-, saliency- and seed-based methods to generate proposals from CAMs/attention maps, respectively. Moreover, these methods also used image group knowledge provided by human. The image set is divided into multiple groups by human and each group contains a common object.
A common pitfall of these approaches is that they are only able to predict proposals for limited object categories, due to the limitation of pre-trained classification models. Moreover, they search object proposals based on CAMs/attention maps, which highlight important regions in images but ignore objects in non-activated regions. Different from these approaches, our method can generate proposal for open categories and cover diverse objects.

\textbf{CLIP Cue.} CLIP \cite{radford2021learning} is an image-text matching model, which contains a visual feature encoder and a textual feature encoder. It was trained with a large number of image-language pairs to embed the visual features and textual features into a unified feature space. Because it shows good ability to recognize diverse language and object categories, its feature space has been used as cues in many applications, such as image generation \cite{patashnik2021styleclip}, image-text retrieval \cite{sun2021lightningdot, bai2021connecting, dzabraev2021mdmmt}, image classification \cite{sariyildiz2021concept, cheng2021data} and image captioning \cite{changpinyo2021conceptual}. Inspired by these works, we exploit CLIP features as cues for unsupervised open-category object proposal generation. 

\section{Proposed Method}

\begin{figure}
	\centerline{\includegraphics[scale=0.55]{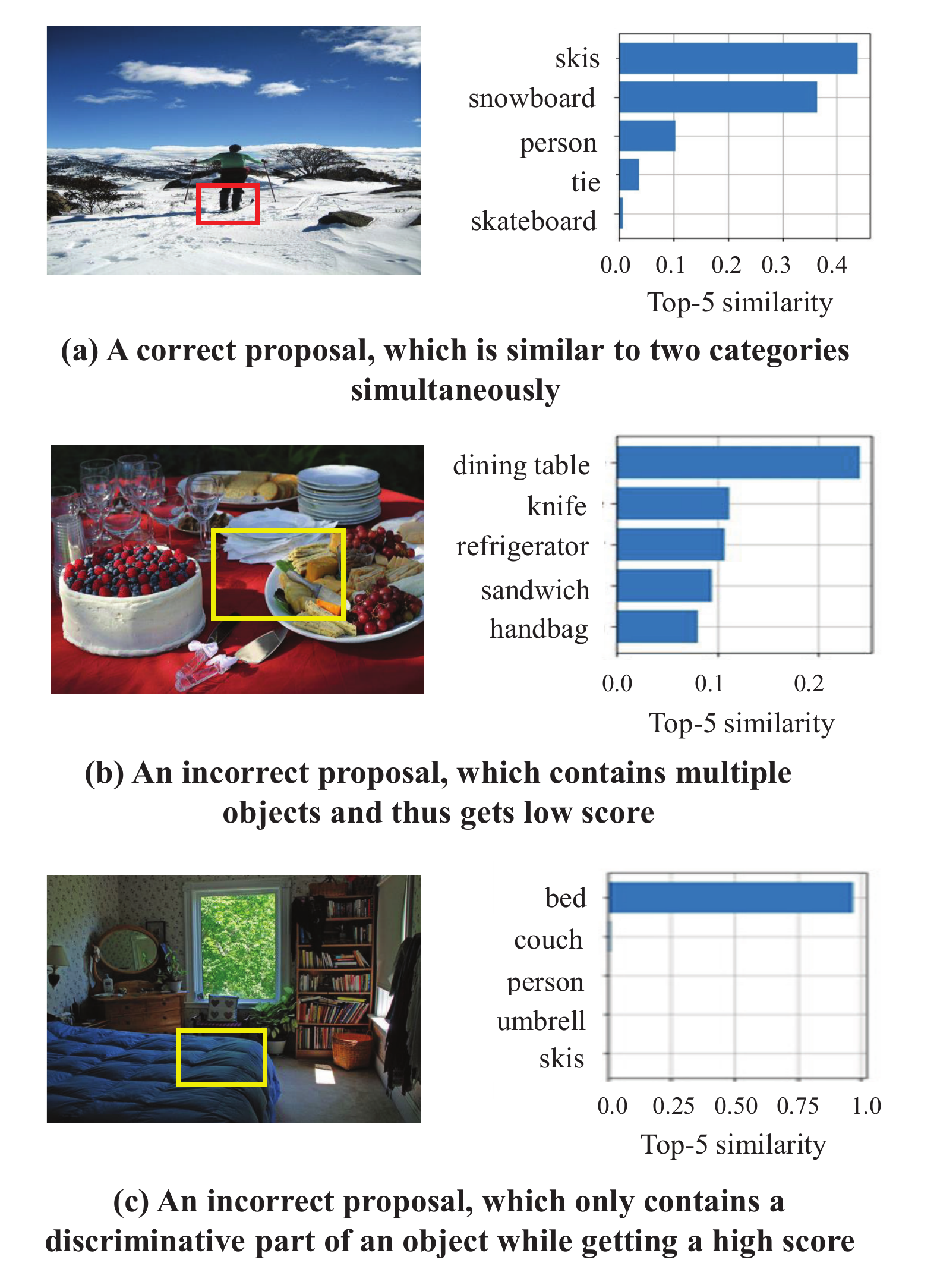}}
	\caption{Some examples of CLIP matching results for proposals.}
	\label{fig_sim}
\end{figure}

\begin{figure*}
	\centerline{\includegraphics[scale=0.35]{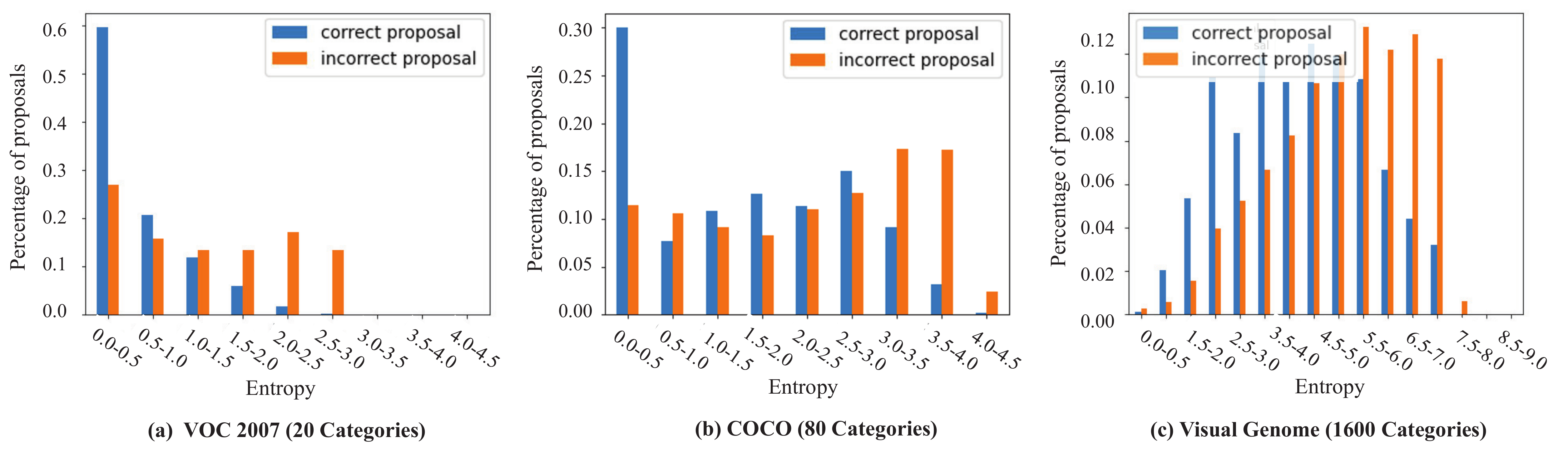}}
	\caption{Distributions of similarity entropies on the PASCAL VOC 2007, COCO and Visual Genome training sets. Correct proposals mean their IoU with ground truths higher than 0.5, while incorrect proposals with IoU lower than 0.5.}
	\label{fig_entropy}
\end{figure*}

\subsection{CLIP Feature Analysis for Proposal Generation}
\label{analysis}
Previous deep-learning-based unsupervised proposal generation methods leverage pre-trained classifiers as prior knowledge to extract salient object regions. However, they are only able to generate proposals for a fixed number of object categories, because their classifiers are trained with a fixed set of classes. In this paper, we exploit prior knowledge from CLIP \cite{radford2021learning} image-text matching as cues. CLIP consists of a visual feature encoder (ViT \cite{dosovitskiy2020image}) and a textual feature encoder (GPT-2 \cite{radford2019language}), and embeds visual and textual features into a same feature space for matching.
Unlike classification models, CLIP is trained to match images with their corresponding natural language descriptions, and thus has potential to recognize diverse objects in real world. However, directly extracting objects from CLIP attention maps like previous methods ignores objects in non-salient regions. It is also hard to separate overlapped instances \cite{simeoni2021localizing}. Meanwhile, since CLIP \cite{radford2021learning} is trained to recognize a single object, it cannot directly encode features for an image containing multiple objects.
Therefore, our basic idea is to use an existing proposal method such as Edge Boxes \cite{zitnick2014edge} to extract a mass of candidate single-object proposals, while leveraging CLIP prior knowledge to evaluate their objectness to select proposals.

To evaluate the objectness, we first extract visual features of each initial proposal using CLIP image encoder, while using the textual feature encoder to capture textual features of the candidate object categories. In a specific dataset such as Microsoft COCO \cite{lin2014microsoft}, its object categories (80 categories for COCO) can serve as the candidate object categories. In real world, we can use a large noun dictionary as candidate object categories. After the feature extraction, for each proposal, we calculate its feature similarity with every candidate category and use softmax function to normalize these similarities. A simple way is to use the \textbf{maximum similarity} as the objectness score, because one well-extracted proposal usually has an exact category. However, as shown in Fig. \ref{fig_sim} (a), we observed that in open-category proposal generation, some well-extracted proposals may be simultaneously assigned to multiple categories, because semantically confusing categories are unavoidable in real world. It is also not always accurate to use multiple top similarities to evaluate the proposal objectness. Therefore, rather than using the maximum similarity, we propose a simple yet effective objectness estimation method based on the \textbf{similarity entropy}.

\begin{table}
	\centering
	\scalebox{0.9}{
		\begin{tabular}{lccc}
			\toprule
			Proposals                  &VOC 2007  & COCO & Visual Genome	\\
			\midrule
			Correct proposals          &0.56      &1.29  & 3.38\\
			Incorrect proposals        &1.32      &2.41  & 5.03\\
			\bottomrule
	\end{tabular}}
	\caption{Average similarity entropies of proposals on PASCAL VOC 2007, COCO and Visual Genome training sets.}
	\label{tab_entropy}
\end{table}

Table \ref{tab_entropy} shows average similarity entropies of different proposals on multiple datasets. Fig. \ref{fig_entropy} depicts the distributions of similarity entropies. Here, we use a traditional proposal method \cite{zitnick2014edge} to generate a large number of proposals and divide them into two types, correct and incorrect proposals. Correct proposals are defined as any proposal whose IoU with ground truths are higher than 0.5, while others are incorrect proposals. From Table \ref{tab_entropy} and Fig. \ref{fig_entropy}, we can observe the followings: (1) correct proposals show significantly lower average CLIP similarity entropies than incorrect proposals on all datasets; (2) correct proposals are mainly distributed in low-entropy ranges, while incorrect proposals dominate high-entropy ranges; and (3) different datasets show distinctly different entropy ranges.

\begin{figure*}
	\centerline{\includegraphics[scale=0.5]{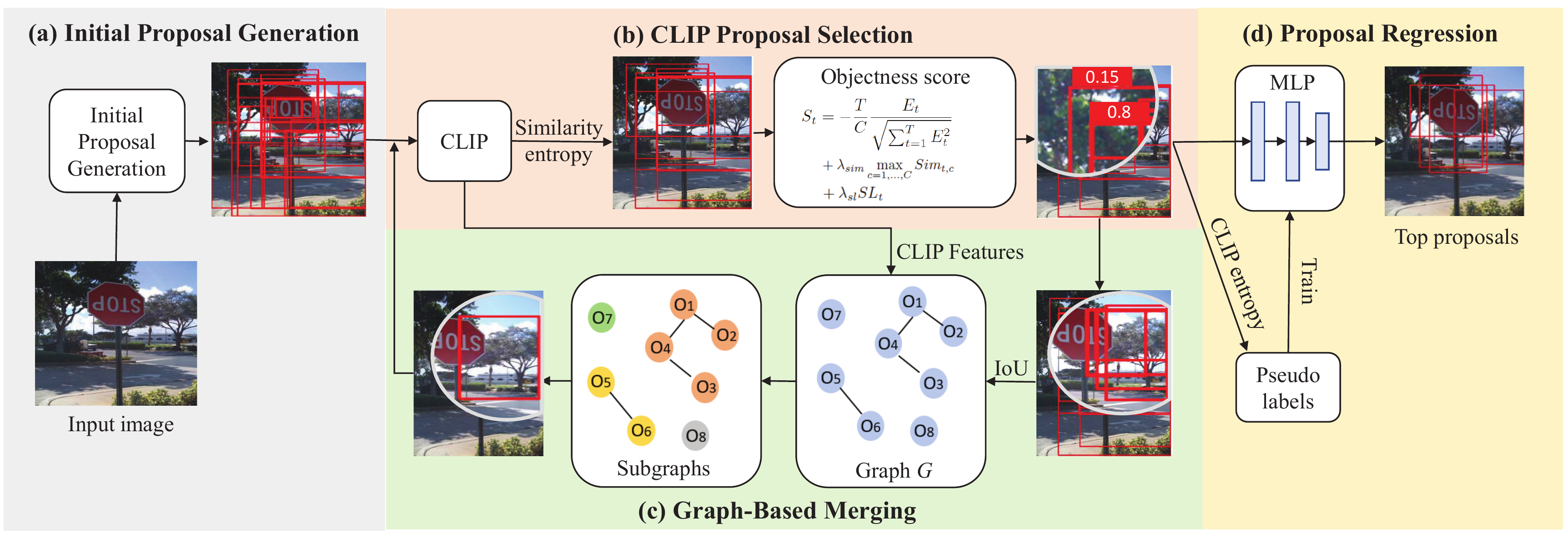}}
	\caption{Illustration of our ProposalCLIP. (a) The initial proposal generation model extracts initial proposals. (b) The CLIP proposal selection model selects and re-scores proposals based on CLIP cues. (c) The graph-based proposal merging model corrects fragmented proposals based on CLIP features. (d) The proposal regression model refines proposals.}
	\label{fig_method}
\end{figure*}

Based on observations (1) and (2), setting a threshold on CLIP similarity entropy can directly filter out about 40\% incorrect proposals while keeping most of the correct proposals. However, as pointed out in the observation (3), different datasets require to carefully set different thresholds, which substantially reduces the generalization ability in real-world applications. Hence, we propose to filter out proposals through the percentage instead of a threshold. For example, we can select 60\% low-similarity-entropy initial proposals to remove a mass of incorrect proposals while keeping correct proposals. Meanwhile, we also propose to use the CLIP similarity entropy to re-score proposals for better generation, as described in Sec. \ref{CLIP Proposal Selection}.

We also observe \textbf{limitations} of the CLIP similarity entropy. As shown in Fig. \ref{fig_sim} (c), for incorrect proposals only containing a discriminative part of an object, CLIP shows low entropies and cannot remove them based on the similarity entropy. To solve this problem, we propose a graph-based proposal merging model in Sec. \ref{Graph-Based Proposal Merging}, which is able to merge such fragmented proposals by leveraging CLIP features. We next introduce our ProposalCLIP based on these observations and analysis. 

\subsection{ProposalCLIP}
Our ProposalCLIP contains four modules, as illustrated in Fig. \ref{fig_method}: (a) an initial proposal generation model to get candidate proposals from the input image, (b) a CLIP proposal selection model that leverages CLIP feature space to refine proposals from the candidate proposals, (c) a graph-based merging model that merges fragmented proposals based on both spatial and CLIP cues, and (d) a proposal regression model that further refines proposals. Below we introduce the details of each module.

\subsubsection{Initial Proposal Generation}
Given the input image $I$, we first obtain initial proposals $\{O_{m} \}_{m=1}^{M}$, where $M$ is the number of initial proposals and $O_{m} \in R^{4}$ represents the coordinates of the $m$-th proposal. Our method can use any existing proposal generation model to predict initial proposals. Here, we use Edge Boxes \cite{zitnick2014edge} as an example, which generates proposals based on low-level image information. Besides proposal coordinates, the existing proposal generation model also predicts an objectness score $SL_{m}$ for each proposal $O_{m}$. This score can also be used in the following selection.

\subsubsection{CLIP Proposal Selection} \label{CLIP Proposal Selection}
Based on the analysis in Sec. \ref{analysis}, we propose to leverage CLIP similarity entropy to estimate objectness of initial proposals and select proposals. Specifically, for each initial proposal $O_{m}$, we first use CLIP vision encoder to extract its features $V_{m}$. Then, we encode textual features $\{T_{c}\}_{c=1}^{C}$ of every candidate object category, where $C$ is the number of candidate categories. Next, we calculate the cosine similarity between the visual features of the proposal and the textual features of each category:
\begin{equation}
	Sim_{m, c} = \frac{V_{m} \cdot T_{c}}{\Vert V_{m} \Vert \Vert T_{c} \Vert} .
\end{equation}
The cosine similarity $Sim_{m, c}$ is then normalized by a softmax function. 

After than, the CLIP similarity entropy for the proposal $O_{m}$ can be obtained as follows:
\begin{equation}
	E_{m} = -\sum_{c=1}^C Sim_{m, c} \times log(Sim_{m, c}) .
\end{equation}
We then filter out high-entropy proposals to remove a large number of incorrect proposals but retain most of correct proposals. The $T$ retained proposals are represented by $\{O_{t} \}_{t=1}^{T}$. In our experiments, we remove 40\% high-entropy proposals.

However, as shown in Fig. \ref{fig_entropy}, there are still many incorrect proposals in the retained proposals. Therefore, we propose a CLIP-based objectness score to re-rank retained proposals for further selection as:
\begin{equation}
	S_{t} = -\frac{T}{C} \frac{E_{t}}{\sqrt{\sum_{t=1}^{T}E_{t}^{2}}} + \lambda_{sim} \max\limits_{c=1,...,C}{Sim_{t, c}} + \lambda_{sl} SL_{t}
\end{equation}
where the first item is the negative entropy score, the second and third items are the maximum similarity and the initial score, respectively. We use them as references, because proposals are also probably correct when they have extremely high maximum similarities and the initial scores. Similarity entropies for different proposals are normalized by L2 normalization. We use  $\frac{T}{C}$ to automatically weight the first item to be in an appropriate range. $\lambda_{sim}$ and $\lambda_{sl}$ are coefficients to control the proportion of each item. We then select out correct proposals based on this objectness score.

\subsubsection{Graph-Based Proposal Merging} \label{Graph-Based Proposal Merging}
Although our CLIP proposal selection model can filter out a large number of incorrect proposals, some fragmented proposals which only contain a discriminative part of an object cannot be easily filtered. Thus, we propose a graph-based proposal merging model to solve this limitation.

Specifically, we first build an undirected graph $\mathcal{G}: \mathcal{G=<N, E>}$. Nodes $\mathcal{N}$ in the graph are proposals $\{O_{t} \}_{t=1}^{T}$ selected by our CLIP proposal selection model. Edges $\mathcal{E}$ are computed by spatial and semantic similarities between these proposals. We use the IoU (intersection over union) between two proposals to evaluate their spatial similarity as follows:
\begin{equation}
	IoU_{i,j} = \frac{O_{i} \cap O_{j}}{O_{i} \cup O_{j}}
\end{equation}
where $IoU_{i,j}$ is the IoU between proposals $O_{i}$ and $O_{j}$. Their semantic similarity $PSim_{i,j}$ is estimated by the cosine similarity between their CLIP visual features:
\begin{equation}
	PSim_{i,j} = \frac{V_{i} \cdot V_{j}}{\Vert V_{i} \Vert \Vert V_{j} \Vert}.
\end{equation}

After capturing both the spatial and semantic similarities, edges in the graph are calculated as:
\begin{equation}
	\begin{split}
	   e_{i,j} & = U(IoU_{i,j} - Thr_{IoU}) \\
  	             & \times U(PSim_{i,j} - Thr_{PSim})
  	\end{split}
\end{equation}
where $e_{i,j} \in \mathcal{E}$ represents the edge between nodes $O_{i}$ and $O_{j}$. $Thr_{IoU}=0.5$ and $Thr_{PSim}=0.9$ are thresholds for IoU and visual feature similarity, respectively. $U(\cdot)$ represents a unit step function. Here, we use a strict criterion to generate edges. There is an edge ($e_{i,j}=1$) between two proposals only when they are well overlapped and have very similar features. 

Once the graph $\mathcal{G}$ is built, we then determine all maximal connected subgraphs $\{\mathcal{H}_{k}\}_{k'=1}^{K'}$. We delete subgraphs which only contain one node (i.e., one proposal) and merge proposals in each of the remaining subgraph $\mathcal{H}_{k'}$. Finally, $K$ merged proposals are generated and represented by $\{\tilde{O}_{k}\}_{k=1}^{K}$.

We use the CLIP proposal selection model to evaluate the objectness of the merged proposals, by generating their similarity entropies $\{\tilde{E}_{k}\}_{k=1}^{K}$ and objectness scores $\{\tilde{S}_{k}\}_{k=1}^{K}$. Some merged proposals are removed, if their entropies are higher than the maximum entropy in selected proposals (i.e., $\tilde{E}_{k} > \max\limits_{t=1,...,T}{E_{t}}$). The rest of the merged proposals are added to the set of selected proposals.

\subsubsection{Proposal Regression}
We further propose a proposal regression model as an optional part to refine proposals. The first three parts in our method do not need any training. If there is an images set without annotations, our proposal regression model can extract pseudo labels from this set and train a regression model for proposal refinement. To extract pseudo labels, we leverage the first three parts in our method to generate proposals. We then select the intersection between top 1\% low-entropy proposals and top 5\% high-initial-score proposals as pseudo labels  $\{Y_{n}\}_{n=1}^{N}$, where $N$ is the number of pseudo labels.

A lightweight MLP (Multilayer Perceptron) is built to regress proposals. Inspired by RPN \cite{ren2015faster}, we take the proposal visual features as the input. We also input the visual features of the entire image and the normalized coordinates of the proposal as references. Our MLP consists of three fully-connected layers with batch normalization and ReLU activation. The first layer is used to fuse input features, the second layer is to transform the fused features and the final layer outputs the normalized coordinates of the refined proposal $\hat{O}$.

The model is trained by Smooth L1 loss as follows:
\begin{equation}
	Loss = \left\{
	\begin{aligned}
		&0.5(\hat{O}_{n} - Y_{n})^{2}, & &\Vert\hat{O}_{n} - Y_{n}\Vert < 1 \\
	    &\Vert\hat{O}_{n} - Y_{n}\Vert - 0.5, & & \Vert\hat{O}_{n} - Y_{n}\Vert > 1
	\end{aligned}\right\}
 \end{equation}
where $\hat{O}_{n}$ is the regressed proposals on the training set and $Y_{n}$ is the pseudo label.

The objectness of refined proposals are also estimated by the CLIP proposal selection model. If the CLIP entropy of the refined proposal is lower than the original one and their IoU is higher than 0.75, we replace the original proposal with the refined one. If not, we keep the original proposal.

\begin{table*}
	\centering
	\scalebox{0.75}{
		\begin{tabular}{l|ccccc|ccccc|ccccc|ccccc}
			\toprule
			&\multicolumn{10}{c|}{VOC 2007}&\multicolumn{10}{c}{COCO} \\
			\hline
			&\multicolumn{5}{c|}{Recall@0.5 (\%)}  &\multicolumn{5}{c|}{AR(\%)} 	&\multicolumn{5}{c|}{Recall@0.5 (\%)}  &\multicolumn{5}{c}{AR(\%)} \\
			Method                                                  &1   &10  &30  &50  &100  &1  &10  &30  &50  &100 &1   &10  &30  &50  &100  &1  &10  &30  &50  &100\\
			\midrule
			\emph{Fully-supervised}                                 &&&&&&&&&&\\
			DeepBox \cite{kuo2015deepbox}                            &- &58.1&71.8&77.2&84.5 &-&33.9&44.5&49.2&54.9 &-&21.9&32.3&38.4&47.5 &-&12.5&18.9&22.5&27.8    \\   
			RPN \cite{ren2015faster}                                &- &60.1&73.8&80.7&89.0 &- &28.4&38.1&42.7&48.9 &-&30.6 &46.2&55.1 &65.0 &-&16.1&25.0&30.2&36.1    \\
			RefinedBox \cite{liu2020refinedbox}                     &- &79.5&88.6&90.8&92.4 &- &49.8&56.1&57.7&59.0 &-&44.6&57.3&62.4&68.1 &-&30.4&38.2&41.1&44.3    \\
			\hline
			\emph{Unsupervised}                                     &&&&&&&&&&\\
			Selective search \cite{uijlings2013selective}          &11.3&35.7&52.3&59.8&69.1 &4.9&16.5 &27.7 &33.9 &42.0 &3.3&11.1&19.6&24.2&31.0&1.5&4.1&8.5&11.2&15.7   \\
			Edge boxes \cite{zitnick2014edge}                      &15.2&42.5&58.3&64.7&72.5 &7.6 &24.2&35.1&39.9&46.3&5.5&17.1&25.9&30.5&36.5&3.0&10.9&16.2&18.7&23.5  \\
			rOSD \cite{vo2020toward}                                &16.6&33.2&42.9&45.3&49.8&6.8&15.3&21.5&22.1&25.5 &4.7 &13.5&22.7&25.4&27.1&1.5&4.6&9.8&12.2&13.9  \\
			LOST \cite{simeoni2021localizing}                       &18.8&23.7&25.4&26.3&27.7 &7.2&10.2&11.7&12.6&13.7 &5.0&6.6&7.4&7.8&8.5 &1.6&2.4&2.8&3.1&3.6   \\
			Ours     &\textbf{22.1}&\textbf{52.1}&\textbf{65.8}&\textbf{71.7}&\textbf{78.0} &\textbf{10.6}&\textbf{29.6}&\textbf{39.3}&\textbf{43.5}&\textbf{48.3}                                               &\textbf{11.2}&\textbf{27.1}&\textbf{33.5}&\textbf{35.7}&\textbf{38.3}&\textbf{4.8}&\textbf{14.3}&\textbf{20.2}&\textbf{23.9}&\textbf{26.8}  \\
			\bottomrule	\end{tabular}}
	\caption{Proposal generation results on the PASCAL VOC 2007 test set and COCO validation set.}
	\label{tab_coco}
\end{table*}

\begin{table}
	\centering
	\scalebox{0.78}{
		\begin{tabular}{l|ccccc}
			\toprule
			                                                        &\multicolumn{5}{c}{Recall@0.5 (\%)} \\
			Method                                                  &1   &10  &30  &50  &100 	\\
			\midrule
			\emph{Cross-domain} \\
			Faster RCNN \cite{ren2015faster} (trained on COCO)     &\textbf{10.3}&22.5&29.0&31.4&32.1\\
			\hline
			\emph{Unsupervised}                                     &&&&&\\
			Selective search \cite{uijlings2013selective}           &4.0&11.3&17.8&28.1&38.7   \\
			Edge boxes \cite{zitnick2014edge}                       &4.9&14.5&23.9&33.8&45.8  \\
			rOSD \cite{vo2020toward}                                &5.6&15.7&25.5&28.3&33.5  \\
			LOST \cite{simeoni2021localizing}                       &5.8&12.1&13.2&13.6&13.9\\
			Ours  &8.5&\textbf{24.1}&\textbf{33.0}&\textbf{38.7}&\textbf{47.1}\\                                                 
			\bottomrule
	\end{tabular}}
    \caption{Proposal generation results on the Visual Genome mini dataset.}
	\label{tab_vg}
\end{table}

\section{Experiments}

\subsection{Datasets and Metrics}
We verify our method on three object proposal generation datasets, PASCAL VOC 2007 \cite{everingham2010pascal}, COCO 2017 \cite{lin2014microsoft} and Visual Genome \cite{krishna2016visual}. PASCAL VOC 2007 \cite{everingham2010pascal} consists of 9,963 images and 20 object categories. It is split into training, validation and test sets containing 2,501, 2,510 and 4,952 images, respectively. Similar to previous works \cite{ren2015faster, liu2020refinedbox}, we verify our method on the test set, while combining images in training and validation sets to train the proposal regression model. COCO 2017 \cite{lin2014microsoft} contains 123,287 images and 80 object categories, which is divided into training set (118,287 images) and validation set (5,000 images). We use the validation set to test, while employing the training set to extract pseudo labels. Visual Genome \cite{krishna2016visual} includes 107,228 images. Since it has over 1,600 object categories, we choose it to evaluate the open-category ability of our method. We randomly select 5,000 images which contain about 50,000 bounding boxes for testing and 2,000 images for training. We call it as ``Visual Genome mini''.

We adopt common proposal generation metrics, Recall and AR (Average Recall), to evaluate the performance. Recall@$X$ is the ratio of well found ground truth objects whose IoU with a proposal is higher than the threshold $X$. AR is the average recall at IoU thresholds from 0.5 to 0.95. We use AP (Average Precision) to evaluate the performance of unsupervised object detection.

\subsection{Implementation Details}
Our method can use any existing proposal generation technique as our initial proposal generation module. In our experiments, we use Edge Boxes \cite{zitnick2014edge} as an example and generate 300 initial proposals per image. In the CLIP proposal selection model , we select 60\% low-entropy proposals and set $\lambda_{sim}$ to 0.06 and $\lambda_{sl}$ to 1. In the graph-based proposal merging model, we set $Thr_{IoU}$ and $Thr_{PSim}$ to 0.5 and 0.9, respectively. In the proposal regression model, we train the MLP for 30 epochs and the learning rate is 1e-5. All experiments are conducted on the Pytorch deep learning platform \cite{paszke2019pytorch} on one Nvidia RTX 3090 GPU.

\subsection{Comparison with State of the Art}
We first compare the unsupervised object proposal generation performance on the VOC 2007 and COCO datasets. The results are shown in Table \ref{tab_coco}. It can be seen that compared with Edge boxes \cite{zitnick2014edge}, our method achieves improvements by a large margin, especially when extracting a small number of proposals (e.g., 1, 10 and 30). When extracting 10 proposals, our method outperforms Edge boxes \cite{zitnick2014edge} by about 10\% on VOC 2007 and COCO, in the term of Recall@0.5. When extracting a large number of proposals, our method also shows significantly improvements. Compared with the CNN-based method rOSD \cite{vo2020toward} and the Transformer-based method LOST \cite{simeoni2021localizing}, our method achieves gains of 3.3\% on VOC 2007 and 6.2\% on COCO in Recall@0.5, when extracting only one proposal. Moreover, rOSD \cite{vo2020toward} and LOST \cite{simeoni2021localizing} are both based on CAMs/attention maps. They do not well generate multiple proposals, while our method can simultaenously predict multiple proposals for an image. We also report the results from some fully-supervised methods, as reported in \cite{liu2020refinedbox}. It can be observed that our method outperforms the fully-supervised method DeepBox \cite{kuo2015deepbox} on COCO, when extracting a small number of proposals. These superior results demonstrate the effectiveness of our method.

We then compare the open-world proposal generation capability on Visual Genome mini \cite{everingham2010pascal}. As shown in Table \ref{tab_vg}, our methods outperforms all previous unsupervised methods. Faster RCNN \cite{ren2015faster} trained on COCO achieves good performance when extracting a few objects (1 and 10 objects), because it is well trained to capture 80-category objects. However, it cannot well extract more objects, due to no annotation for other object categories. Our method shows better recalls when extracting more objects.
Some qualitative results are shown in Fig. \ref{fig_results}. It is can be seen that Faster RCNN \cite{ren2015faster} trained on COCO fails to capture some objects, such as the ``potato'' and ``drawing''. This is because Faster RCNN \cite{ren2015faster} requires annotations for these objects. Our method can recognize open-category objects without annotations, and thus successfully extracts these objects. It is also seen that the top 100 proposals generated by our ProposalCLIP are better than those from Edge boxes \cite{zitnick2014edge}. 

\subsection{Ablation Study} 

\begin{figure*}
	\centerline{\includegraphics[scale=0.47]{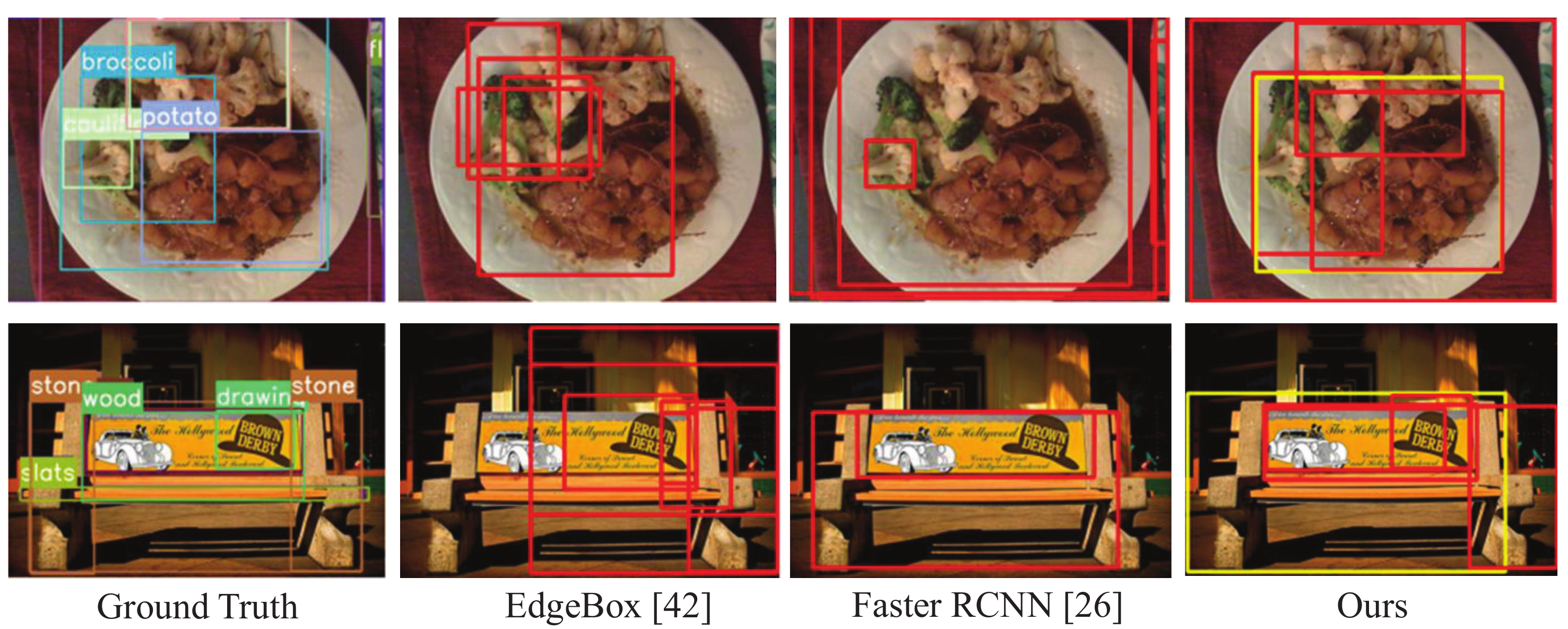}}
	\caption{Qualitative results on the Visual Genome mini dataset. Top to bottom: ground truth, correct proposals from Edge boxes \cite{zitnick2014edge}, Faster RCNN \cite{ren2015faster} trained on COCO and our ProposalCLIP with 100 proposals. Yellow boxes in our results are generated by our graph-based merging model.}
	\label{fig_results}
\end{figure*}

\textbf{The contribution of each component.} Table \ref{tab_as} shows the effects of different components in our method. We find that our CLIP proposal selection model achieves the most improvements while only slightly reducing the speed. The graph-based proposal merging also significantly improves the performance. We visualized proposals generated by our graph-based merging model in Fig. \ref{fig_results}. It can be found that our model successfully captures some objects by the graph-based merging. The proposal regression model decreases the recall when generating a few proposals. This could be caused by unavoidable noises in pseudo labels. Nevertheless, the proposal regression model increases the performance in most cases. These results demonstrate the effectiveness of our CLIP proposal selection, graph-based proposal merging and proposal regression models.

\begin{table}
	\centering
	\scalebox{0.77}{
		\begin{tabular}{l|cccc|c}
			\toprule
			&\multicolumn{4}{c|}{Recall@0.5 (\%)} &\\
			Method                                                  &1   &10  &30  &50 &Time (s)\\
			\midrule
			Initial proposals (Edge boxes \cite{zitnick2014edge})   &15.2&42.5&58.3&64.7 & 0.9\\
			+ CLIP Proposal Selection                               &21.0&51.8&62.1&68.5 & 1.2\\
			+ Graph-Based Proposal Merging                          &21.6&\textbf{52.4}&64.6&70.8& 1.5\\
			+ Proposal Regression                                   &\textbf{22.1}&52.1&\textbf{65.8}&\textbf{71.7} &1.9\\
			\bottomrule
	\end{tabular}}
	\caption{The effects of different components on the PASCAL VOC 2007 test set. Time means the running time per image.}
	\label{tab_as}
\end{table}

\begin{table}
	\centering
	\scalebox{0.79}{
		\begin{tabular}{l|ccccc}
			\toprule
		                                  & \multicolumn{5}{c}{Recall@0.5 (\%)}\\
			Objectness score                &1   &10  &30  &50  &100\\
			\midrule
			Edge boxes \cite{zitnick2014edge}  &15.2&42.5&58.3&64.7&72.5\\
			\hline
			Initial score                 &19.1&46.5&61.2&65.7&74.4\\
			Maximum similarity                  &17.6&50.7&60.0&65.3&74.1\\
			CLIP similarity entropy          &20.2&\textbf{52.5}&60.4&66.6&75.0\\
			Our final objectness score  &\textbf{21.0}&51.8&\textbf{62.1}&\textbf{68.5}&\textbf{76.3}\\
			\bottomrule
	\end{tabular}}
	\caption{The effects of different objectness scores on the PASCAL VOC 2007 test set.}
	\label{tab_score}
\end{table}

\textbf{Objectness scores.} We report the effects of different objectness scores in Table \ref{tab_score}. It can been seen that our similarity entropy outperforms the initial score and maximum similarity. The initial score gains better performance than the original Edge boxes \cite{zitnick2014edge}, thanks to our CLIP entropy selection. Our final objectness score achieves the best performance. 

\textbf{ProposalCLIP for downstream tasks.} We further conduct an experiment to demonstrate the usefulness of our proposals for downstream tasks. Table \ref{tab_task} presents the performance on unsupervised object detection on COCO. We use CLIP \cite{radford2021learning} as a classifier to classify each proposal and leverage NMS (Non-Maximum Suppression) to generate the final results. It can be observed that our Proposal CLIP achieves an 8.5\% AP in an unsupervised manner.

\begin{table}
	\centering
	\scalebox{0.8}{
		\begin{tabular}{l|c}
			\toprule
			Method                             &AP@0.5 (\%)  	\\
			\midrule
			Edge boxes \cite{zitnick2014edge} + CLIP \cite{radford2021learning} + NMS  &6.3\\
			LOST \cite{simeoni2021localizing} + CLIP \cite{radford2021learning} + NMS  &5.2\\
			Ours + CLIP \cite{radford2021learning} + NMS                   &\textbf{8.5} \\
			\bottomrule
	\end{tabular}}
	\caption{Unsuperivsed object detection results on the COCO validation set.}
	\label{tab_task}
\end{table}

\textbf{Limitations.} Our method cannot well deal with small objects, such as the ``cauliflower'' object in the first image in Fig. \ref{fig_results}. Because small objects are usually in low resolution, it is hard to recognize them using CLIP. Meanwhile, the initial proposal generation model is also hard to capture small objects. We leave the exploration of super-resolution techniques to solve this problem for future work.

\section{Conclusion}
In this paper, we have presented ProposalCLIP, an effective approach for unsupervised open-category object proposal generation. In our approach, a proposal selection model is first introduced to recognize open-category objects and select proposals for them by CLIP-based objectness scores. Secondly, a graph-based merging model unifies fragmented proposals based on CLIP feature cues. Thirdly, we introduce a regressor module that leverages CLIP cues to refine proposals. Experimental results have demonstrated that our proposed method is capable to find open-category proposals in an unsupervised manner. Our method also outperforms existing state-of-the-art methods by a large margin on three popular datasets, and shows the benefits for downstream tasks.

\textbf{Social impacts}: We test our method on three popular datasets, which might contain ethic biases. It would be better to collect more unbiased data in real world to reduce this impact.

{\small
\bibliographystyle{ieee_fullname}
\bibliography{my_reference}
}

\end{document}